\documentclass[letterpaper, 10 pt, conference]{ieeeconf}  

\IEEEoverridecommandlockouts                              

\usepackage{graphicx}
\usepackage{amsmath}
\usepackage{amssymb}
\usepackage{stfloats}

\usepackage{xcolor}
\usepackage{url}

\usepackage{graphicx}
\usepackage{amsmath}
\usepackage{amssymb}
\usepackage{stfloats}
\usepackage{url}
\usepackage{tensor}
\usepackage{mathtools}
\usepackage{amsmath}
\usepackage{bbm}
\usepackage{enumerate}
\usepackage[ruled,linesnumbered]{algorithm2e}
\usepackage{siunitx}
\usepackage{tabularx}
\usepackage{multirow}

\usepackage{booktabs}
\usepackage{makecell}
\usepackage{amsmath}
\usepackage[ruled,linesnumbered]{algorithm2e}

\usepackage{tabularx}
\usepackage{multirow}
\usepackage{cite}
\DeclareSymbolFont{largesymbol}{OMX}{yhex}{m}{n}
\DeclareMathAccent{\Widehat}{\mathord}{largesymbol}{"62}

\title{\LARGE \bf Style Transfer Enabled Sim2Real Framework for Efficient Learning of Robotic Ultrasound Image Analysis Using Simulated Data
}

\author{Keyu~Li,
        Xinyu~Mao,
        Chengwei~Ye,
        Ang~Li,
        Yangxin~Xu,
        and~Max~Q.-H.~Meng$^*$,~\IEEEmembership{Fellow,~IEEE}
\thanks{This work was partially supported by National Key R\&D program of China with Grant No. 2019YFB1312400, Hong Kong RGC CRF grant C4063-18G, Hong Kong RGC GRF grant \#14211420 and Hong Kong RGC TRS grant T42-409/18-R awarded to Max Q.-H. Meng.}
\thanks{K. Li, X. Mao, C. Ye, A. Li and Y. Xu are with the Department of Electronic Engineering, The Chinese University of Hong Kong, Hong Kong SAR, China (e-mail: kyli@link.cuhk.edu.hk; maoxinyu@link.cuhk.edu.hk; cwye@link.cuhk.edu.hk; psw.liang@link.cuhk.edu.hk; yxxu@link.cuhk.edu.hk).}
\thanks{
Max Q.-H. Meng is with Shenzhen Key Laboratory of Robotics Perception and Intelligence and the Department of Electronic and Electrical Engineering at Southern University of Science and Technology in Shenzhen, China. He is a Professor Emeritus in the Department of Electronic Engineering at The Chinese University of Hong Kong in Hong Kong and was a Professor in the Department of Electrical and Computer Engineering at the University of Alberta in Canada. (e-mail: max.meng@ieee.org).
}
\thanks{$^*$Corresponding author.}
}

\begin{document}

\maketitle
\thispagestyle{empty}
\pagestyle{empty}

\begin{abstract}
Robotic ultrasound (US) systems have shown great potential to make US examinations easier and more accurate. Recently, various machine learning techniques have been proposed to realize automatic US image interpretation for robotic US acquisition tasks. However, obtaining large amounts of real US imaging data for training is usually expensive or even unfeasible in some clinical applications. An alternative is to build a simulator to generate synthetic US data for training, but the differences between simulated and real US images may result in poor model performance.
This work presents a Sim2Real framework to efficiently learn robotic US image analysis tasks based only on simulated data for real-world deployment. A style transfer module is proposed based on unsupervised contrastive learning and used as a preprocessing step to convert the real US images into the simulation style. Thereafter, a task-relevant model is designed to combine CNNs with vision transformers to generate the task-dependent prediction with improved generalization ability. 
We demonstrate the effectiveness of our method in an image regression task to predict the probe position based on US images in robotic transesophageal echocardiography (TEE). Our results show that using only simulated US data and a small amount of unlabelled real data for training, our method can achieve comparable performance to semi-supervised and fully supervised learning methods. Moreover, the effectiveness of our previously proposed CT-based US image simulation method is also indirectly confirmed. 
\end{abstract}


\section{Introduction}
With the continuous advancement of robotic technology and machine intelligence, robot-assisted ultrasound (US) image acquisition has become a fast-evolving research direction that can assist doctors in performing more efficient and accurate US examinations, while reducing the examination time and operator dependence \cite{li2021overview}. 
Existing studies have proposed a variety of scanning path planning methods for robotic US acquisition, including methods based on registration between pre-operative tomographic images and intraoperative US images \cite{Hennersperger2016MRI, wang2016robotic}, skin surface information \cite{huang2018fully, ning2021autonomic}, and real-time analysis of US images \cite{jiang2020autonomous, li2021image, li2023tee}. Unlike the former two methods, the latter based on US image analysis does not require preoperative manual planning or additional installation of patient tracking devices, thus being widely applicable to both extracorporeal and intracorporeal US imaging applications \cite{li2023tee}.
These methods plan the movement of the probe based on task-relevant hand-crafted features detected in US images \cite{nakadate2010implementation} or features learned via deep learning \cite{jiang2020autonomous}. While the learning-based methods can directly learn high-level representation of images from raw data and generalize to different applications, these methods usually require a large amount of high-quality labelled data to achieve good performance \cite{aiusanalysis}. However, the acquisition of real US data is often time-consuming and labour-intensive, and sometimes even impracticable in certain applications such as semi-invasive intracorporeal US imaging (e.g., transesophageal echocardiography, TEE).
Therefore, it is a promising alternative to build a realistic US simulation environment to train learning-based algorithms for robotic US image analysis and probe navigation tasks.

Existing methods for US simulation can be divided into three categories. The first group of methods mainly build the simulation environment based on real US images. Milletari \textit{et al.} \cite{milletari2019straight} and Hase \textit{et al.} \cite{hase2020ultrasound} used a set of spatially tracked US images acquired on a grid over the patient to construct a discretized virtual patient. Li \textit{et al.} proposed to reconstruct volumetric data based on 2D image sequences so that any slice with arbitrary position and orientation can be sampled in the volume, which can significantly enlarge the state-action spaces and realistically reproduce the probe-tissue interaction in US examinations \cite{li2021autonomous, li2021virtual, li2021image}. The second category uses artificially synthesized images to simulate the structure of specific organs. For instance, Bi \textit{et al.} constructed tube-like 3D volumetric data with binary values to simulate the vessels, and trained a reinforcement learning (RL) agent to search for the longitudinal section along the vessel \cite{bi2022vesnet}. However, such simulation methods are only applicable to organs with relatively simple structures. The final category uses tomographic images (e.g., CT) to simulate the US images based on the US imaging physics \cite{piorkowski2012transesophageal, velikova2022cactuss, li2023enabling}. For instance, we built a TEE simulator based on the CT data of real subjects to train an RL agent for TEE probe guidance in the esophagus to acquire different standard views of the heart \cite{li2023tee}. The CT-based US simulation can better preserve the anatomical features of real human organs and take into account patient diversity in the simulation. However, the differences between simulated and real US images may hinder the performance of the learning-based methods in real-world applications. So far, our simulation-based probe navigation method has only been validated in simulation \cite{li2023tee}, and the simulator itself has been qualitatively demonstrated in real-world US-guided spine procedures \cite{li2023enabling}. 

\begin{figure*}[t]
\setlength{\abovecaptionskip}{-0.0cm}
\centering
\includegraphics[scale=1.0,angle=0,width=0.99\textwidth]{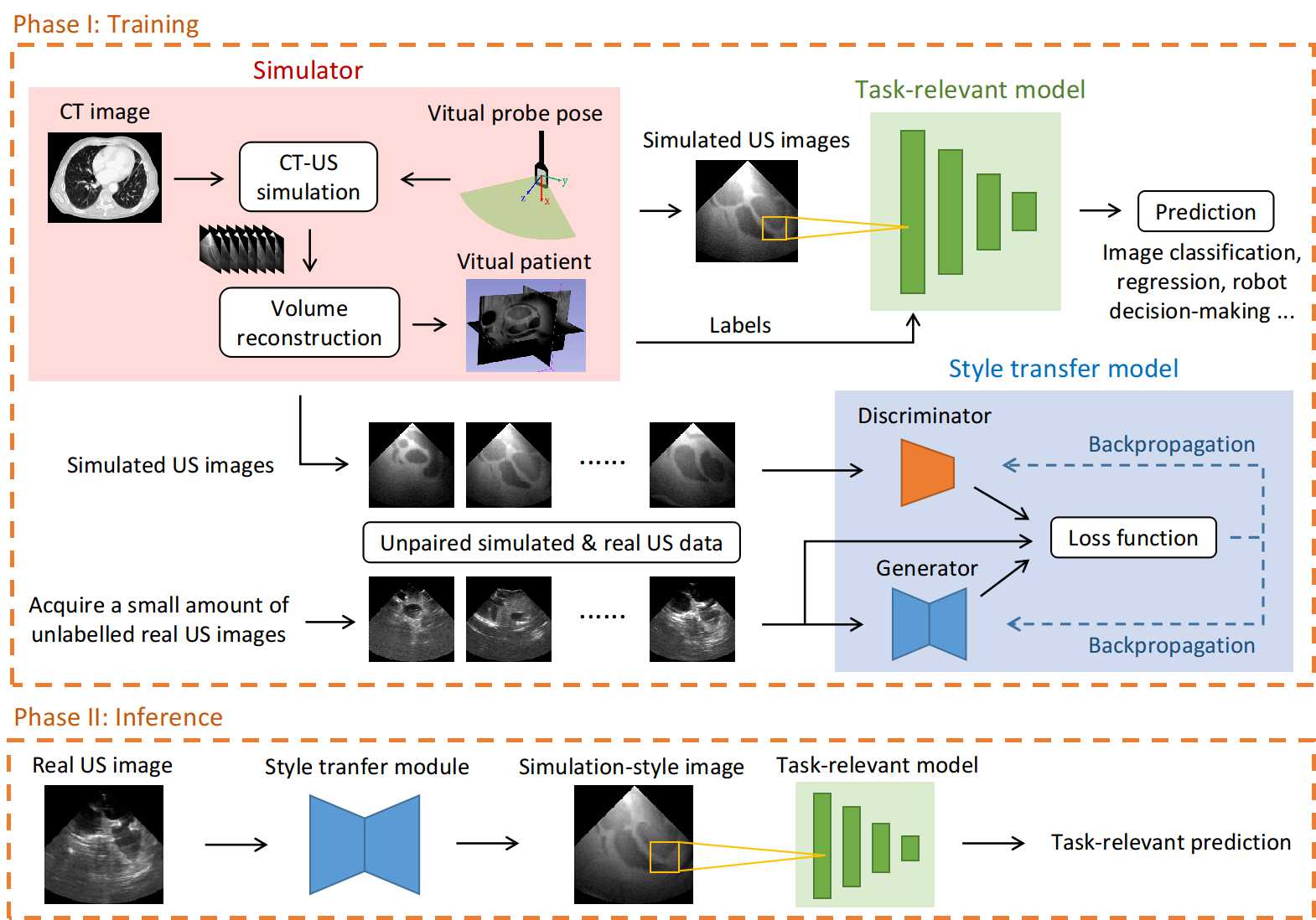}
\caption{(a) Illustration of the Sim2Real framework. During training, the simulator (red) generates simulated US data based on preoperative CT images. We first performed CT-US simulation to generate simulated US images of the target organ, and then conducted volume reconstruction to construct a virtual patient. Then, simulated US images with labels are generated from the virtual patient for the learning of the task-relevant model (green). Meanwhile, the simulated images and a small amount of unlabelled real US images are used to train a style translation model based on unsupervised contrastive learning (blue). During inference, the real US images are first converted via the style transfer module to generate simulation-style images before final prediction is made by the task-relevant model. }
\label{Fig_framework}
\end{figure*}

In order to address the simulation-reality gap in robotic US image analysis and quantitatively demonstrate the effectiveness of our simulation method in real-world applications, in this paper, we propose a Sim2Real framework that combines style transfer and self-attention techniques to efficiently learn robotic US image analysis tasks based only on simulated data for the application on real US images. As shown in Fig. \ref{Fig_framework} The proposed method only uses simulated US data to train the task-relevant model, and uses simulated data and a small amount of unlabeled real US data to train a style transfer model based on contrastive learning. During inference, real images are converted via the style transfer model to generate simulation-style images, which are fed into the task-relevant model as input. We validate the proposed framework in the task of US image-based probe localization in TEE.  This result also indirectly proves the effectiveness of our proposed US image simulator based on CT images.

The contributions of this paper can be summarized as follows:

\begin{itemize}
\item  We propose a Sim2Real framework for efficient learning of robotic US image analysis tasks based on simulated data for deployment in real-world applications. A style transfer module is designed based on unsupervised contrastive learning to combine the content of the real images and the style of the simulated images, and is adopted as a preprocessing step in our proposed framework.
\item Furthermore, we show that using a combination of CNNs and self-attention mechanisms can improve the generalization ability of the task-relevant model to better bridge the gap between simulated and real US images.
\item  Our experimental results in an image regression task show that without using labeled real US data for training, the proposed Sim2Real framework can achieve comparable performance to semi-supervised and fully supervised methods.
\item The effectiveness of our proposed CT-based US image simulator \cite{li2023enabling} is indirectly confirmed, which suggests that the proposed framework has the potential to circumvent the need for obtaining large-scale real data for robotic US image analysis tasks in clinical settings.
\end{itemize}

The rest of this paper is organized as follows. In Section~II, we will introduce the methodology, including the problem formulation, and the design of the task-relevant model and the style transfer module. In Section~III, we will present and discuss the experimental results. Finally, we draw some conclusions in Section~IV.

\section{Methodology}

\begin{figure}[t]
\setlength{\abovecaptionskip}{-0.0cm}
\centering
\includegraphics[scale=1.0,angle=0,width=0.38\textwidth]{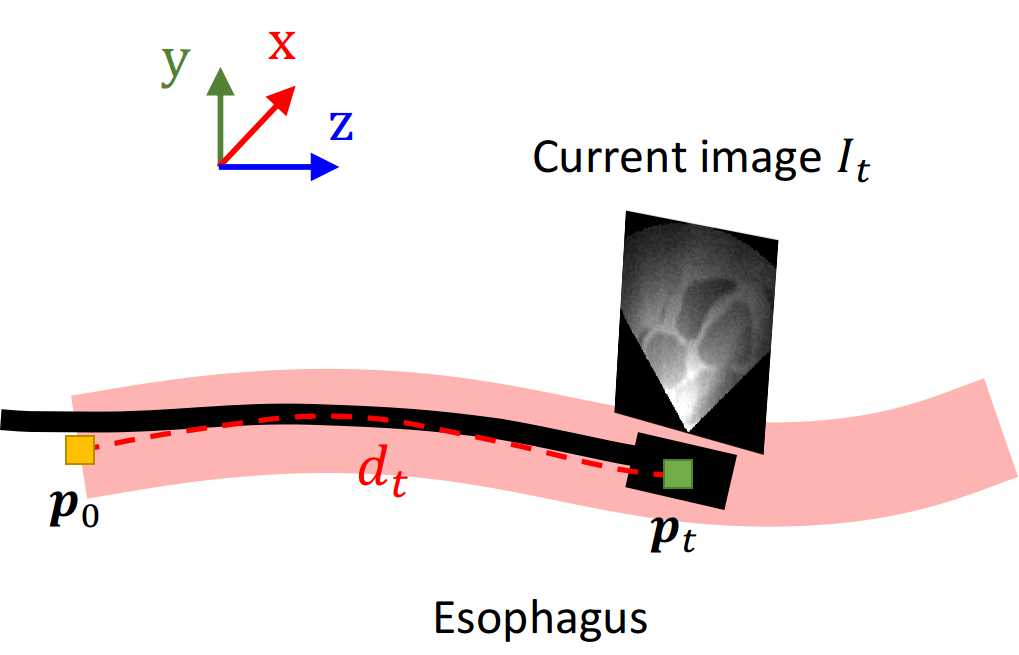}
\caption{Illustration of problem formulation in this work. The starting point and current position of the probe are represented by the yellow and green squares, respectively. The current US image acquired by the probe is shown in green. The depth of the probe is calculated along the centerline of the esophagus (red dashed line). }
\label{Fig_problem}
\end{figure}

\subsection{Problem Formulation}

In this work, we focus on a straightforward task of predicting the probe position from US images in TEE to build and demonstrate our proposed Sim2Real framework. 

Assume that the world coordinate frame is defined as shown in Fig. \ref{Fig_problem} such that the esophagus is monotonic along the $z$-direction. At any time $t$, presumed that the position of the probe in the world coordinate frame can be obtained by a tracking system and is denoted as $\mathbf{p}_t$. The US image collected by the probe corresponding to the current position is denoted as $I_t$. For simplicity and generality of results, we further represent the position of the probe as its depth in the esophagus $d_t$, which is calculated by

\begin{equation}
\label{pos_distance}
d_t = \int_{\mathbf{p}_0}^{\mathbf{p}_t} \! \mathrm{d}l = \int_{z_0}^{z_t} \! |\mathbf{e^{\prime}}(z)| \, \mathrm{d}z
\end{equation}

\noindent where $z_0$ and $z_t$ are the $z$-coordinates of the starting point of the esophagus $\mathbf{p}_0$ and the projection of the current probe position $\mathbf{p}_t$ on the centerline of the esophagus, respectively. $\mathbf{e^{\prime}}(z)$ is the derivative of the esophagus curve. 

The task is to find the mapping function to predict the probe depth $d_t$ based on the US image $I_t$ at any time $t$, i,e., $f\colon {I}_t \mapsto {d}_t$.

\subsection{Task-Relevant Model using CNN-ViT Structure}
Recently, vision transformer (ViT) has become a hot spot in the research field of CV due to its remarkable ability to capture long-range dependencies in data and extract global information from images thanks to the self-attention mechanism \cite{han2022survey}. Prior arts have shown the superiority of ViTs over CNNs in terms of generalization ability for medical image analysis tasks \cite{he2022transformers, li2023tee}. Therefore, it is reasonable to deduce that a combination of CNNs and ViTs may better bridge the gap between the simulated and real US images in our targeted application.

\begin{figure*}[t]
\setlength{\abovecaptionskip}{-0.0cm}
\centering
\includegraphics[scale=1.0,angle=0,width=0.98\textwidth]{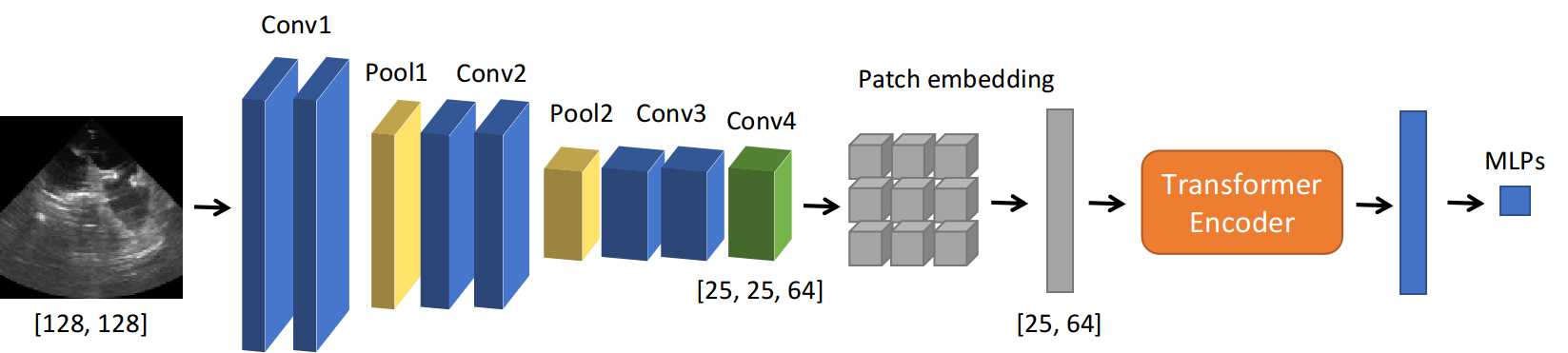}
\caption{Illustration of the design of our \textit{TEEDepthNet}, which is the task-relevant model in this work. The input grayscale image is resized to (128, 128) and processed by four consecutive CNN blocks for high-level feature extraction. The convolutional layers in Conv1, Conv2 and Conv3 all use a kernel size of 3 and a stride of 1, and each is followed by batch normalization. The max pooling layers Pool1 and Pool2 use a kernel size of 2. The Conv3 block is followed by a convolutional layer with a kernel size of 1 and a stride of 1 and a batch normalization layer to convert the feature map to the size of (25, 25, 64), and then embedded as 25 patches with a projection dimension of 64. The ViT encoder has 3 transformer layers, each using 4 heads in the multi-head self-attention module and 128 hidden units in the multi-layer perception (MLP) heads. The final MLP output the predicted depth of the TEE probe in the esophagus. }
\label{Fig_network}
\end{figure*}

In this work, we design a lightweight task-relevant model composed of concatenated CNNs and a ViT encoder for the aforementioned image regression task. We name the task-relevant model as \textit{TEEDepthNet}. As shown in Fig. \ref{Fig_network}, the input US image is fed to four consecutive CNN blocks for high-level feature extraction. The output feature map of the CNN blocks is then embedded as 25 patches with a projection dimension of 64 to fed into a ViT encoder. It has 3 transformer layers, each using 4 heads in the multi-head self-attention module and 128 hidden units in the multi-layer perception (MLP) heads. The final prediction of the TEE probe depth in the esophagus is given by the output of the MLPs. Our implementation of ViT is based on \cite{dosovitskiy2020image}.

\subsection{Style Transfer Module based on Unsupervised Contrastive Learning}

Given a task-relevant model trained only with simulated data, we hope that the style transfer (ST) module, which can be viewed as a preprocessing module for real US images, can convert the input real image to take on the style of the simulated US images, while preserving the structure of the original image. 

The CUT method  \cite{park2020contrastive} is an unsupervised style transfer approach based on contrastive learning, which can be applied on unpaired data across domains and can even be trained on single images. Since we assume that the amount of real US data is very small and no labels are available, CUT is very appropriate for the intended Sim2Real transfer task. Therefore, in this work, we design the ST module based on the CUT method with some modifications to enhance content consistency for the targeted application.

Denote the input domain as $\mathcal{R}$ (represents ``real") and the output domain as $\mathcal{S}$ (represents ``simulation"). Given an unpaired dataset $R=\{I_R \in \mathcal{R}\}$, $S=\{I_S \in \mathcal{S}\}$, we want to convert any image from the input domain $I_R \in \mathcal{R}$ to appear similar to an image from the output domain $I_S \in \mathcal{S}$, while maintaining the content of the input image. The CUT method uses an encoder followed by a decoder as the generator function $G$ (see the blue block in Fig. \ref{Fig_framework}) to generate the output image $\widehat{I_S}$. The original loss function in the CUT method includes an adversarial loss $\mathcal{L}_{GAN}$ \cite{goodfellow2020generative} to enforce the similarity between the output and images from the target domain, and a patchwise contrastive loss $\mathcal{L}_{PatchNCE, R}$ is computed based on noise contrastive estimation (NCE) \cite{oord2018representation} to maximize the mutual information between the input and output. It also includes a patchwise contrastive loss $\mathcal{L}_{PatchNCE, S}$ computed on images from domain $\mathcal{S}$ as a regularizator to prevent unnecessary change of the generator. In our implementation, we modify the loss function in the original CUT method by explicitly taking into consideration the structural similarity (SSIM) \cite{wang2004image} between the original and style transferred images, in order to better preserve the structural information of the real image. This is important for our intended robotic US acquisition application that heavily relies on accurate localization of anatomical structures in the image for the subsequent decision-making tasks \cite{li2021autonomous}. The overall loss function is formulated as:

\begin{equation} \small
\mathcal{L} = \mathcal{L}_{GAN} + \lambda_1 \mathcal{L}_{PatchNCE, R} + \lambda_2 \mathcal{L}_{PatchNCE, S} + \lambda_3 \mathcal{S}(G, R)
\end{equation}

\noindent where $\mathcal{S}(G, R)$ is the SSIM between the generator output and the input image. Based on a number of trials, we empirically select the hyper-parameters $\lambda_1=5$, $\lambda_2= 1$, and $\lambda_3=-1$ in our ST module to yield a good performance.

\section{Experimental Results} \label{simulation}

\subsection{Experimental Setup}

\begin{figure*}[t]
\setlength{\abovecaptionskip}{-0.0cm}
\centering
\includegraphics[scale=1.0,angle=0,width=0.99\textwidth]{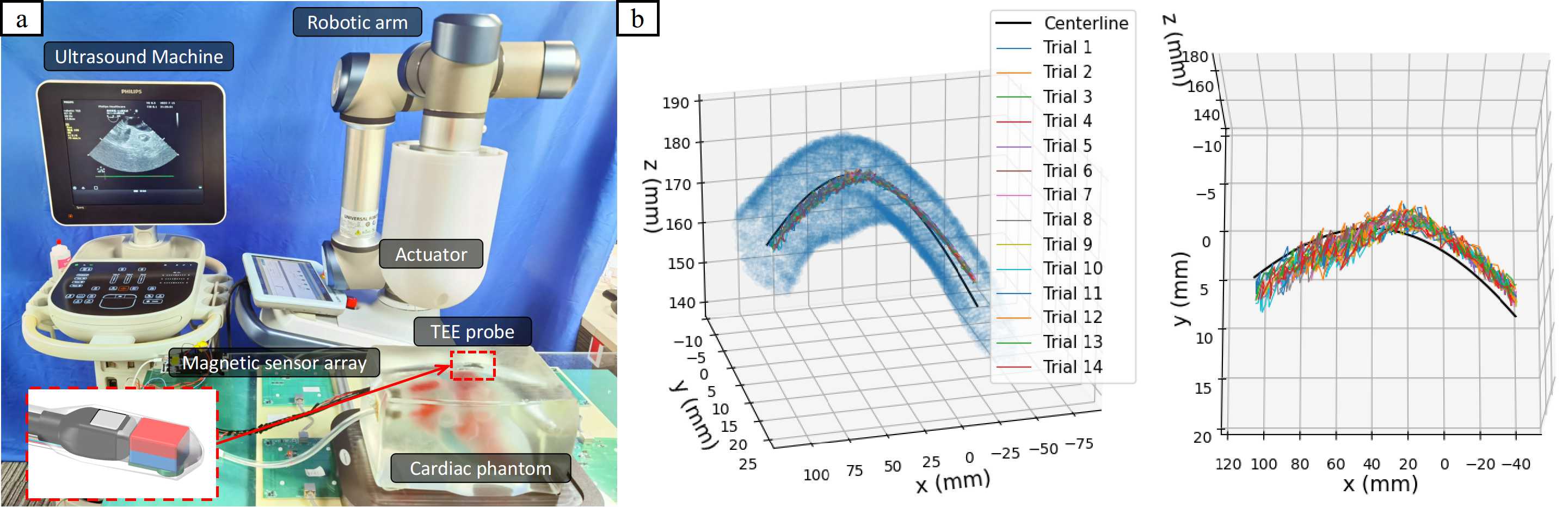}
\caption{(a) The magnetic manipulation system for robotic TEE acquisition, regenerated from \cite{li2023closedloop}. The TEE probe is attached to a permanent magnet and an IMU sensor and placed in the esophagus of a cardiac US imaging phantom, and the actuator containing a permanent magnet is moved above the phantom to control the 5-DOF pose of the probe based on magnetic interaction. The 6D pose of the probe tip is estimated in real time based on the fusion of IMU data and external magnetic field sensor readings. (b) 3D visualization of the manually segmented esophageal points in the phantom (blue), esophagus centerline (black), and 14 real probe trajectories (rainbow) collected using the robotic manipulation system in (a).}
\label{Fig_setup}
\end{figure*}

\subsubsection{Dataset}
In this work, we used a realistic echocardiography training phantom (Blue Phantom Cardiac Echo Model BPH700, CAE Healthcare) to generate the simulated and real TEE datasets for an experimental validation of the proposed Sim2Real framework. To collect real US images and the corresponding probe positions, we used a magnetic manipulation system presented in \cite{li2023closedloop} to estimate the probe poses as the probe is controlled to move along the esophagus in the phantom, as shown in Fig. \ref{Fig_setup}(a). The probe positions are estimated by the localization system based on external and internal sensor fusion with an accuracy of approximately \SI{2}{\mm} and \ang{3} \cite{li2023closedloop, li2022external}. Finally, 14 robotic acquisition trajectories were obtained, comprising a total of 1891 US images and corresponding probe poses. 

To build the TEE simulator for training of the ST module and the task-relevant model, we first performed a CT scan of the phantom, and the esophageal points were manually segmented by a medical expert. Then, the centerline of the esophagus was calculated based on polynomial fitting and B-spline interpolation, as shown in Fig. \ref{Fig_setup}(b-c). By sampling virtual probe poses along the esophagus, we generated a set of spatially tracked synthetic US images using the CT-US simulation method as presented in \cite{li2023tee, li2023enabling}, as shown in Fig. \ref{Fig_framework}. Thereafter, we transferred the recorded probe poses during the robotic acquisition to the TEE simulator to collect 14 corresponding trajectories (with 1891 image-depth pairs) in simulation.

\subsubsection{Training details}
We randomly selected 4 trajectories collected in the real-world acquisitions that contain a total 563 image-depth pairs for testing of the framework, and the remaining 10 trajectories consisting of 1328 image-depth pairs in both simulated and real datasets are used for training purposes. 

For the training of the task-relevant model, three configurations of training datasets are employed: i) \textit{real data}: using 10 real trajectories for training, ii) \textit{combined data}: using 10 simulated trajectories and one real trajectory for training, and iii) \textit{simulation data}: only using 10 simulated trajectories for training. We trained the ``CNN + ViT" model for 100 epochs with a batch size of 16 using Adam optimizer. The initial learning rate is \num{1e-5}, which is decayed by 0.1 once the number of epoch reaches 30, 50, and 80. For comparison, we also trained a CNN-only regression model with the same feature extraction block as Fig. \ref{Fig_network} and attached with a CNN top layer for 100 epochs with an initial learning rate of 0.1. For the training of the unsupervised ST module, one trajectory out of the 10 trajectories containing 141 images in both the simulated and real settings are used to train the ST model for 400 epochs with a learning rate of \num{1e-6}. 

\subsection{Evaluation of the Style Transfer Module}
\begin{figure}[t]
\setlength{\abovecaptionskip}{-0.0cm}
\centering
\includegraphics[scale=1.0,angle=0,width=0.49\textwidth]{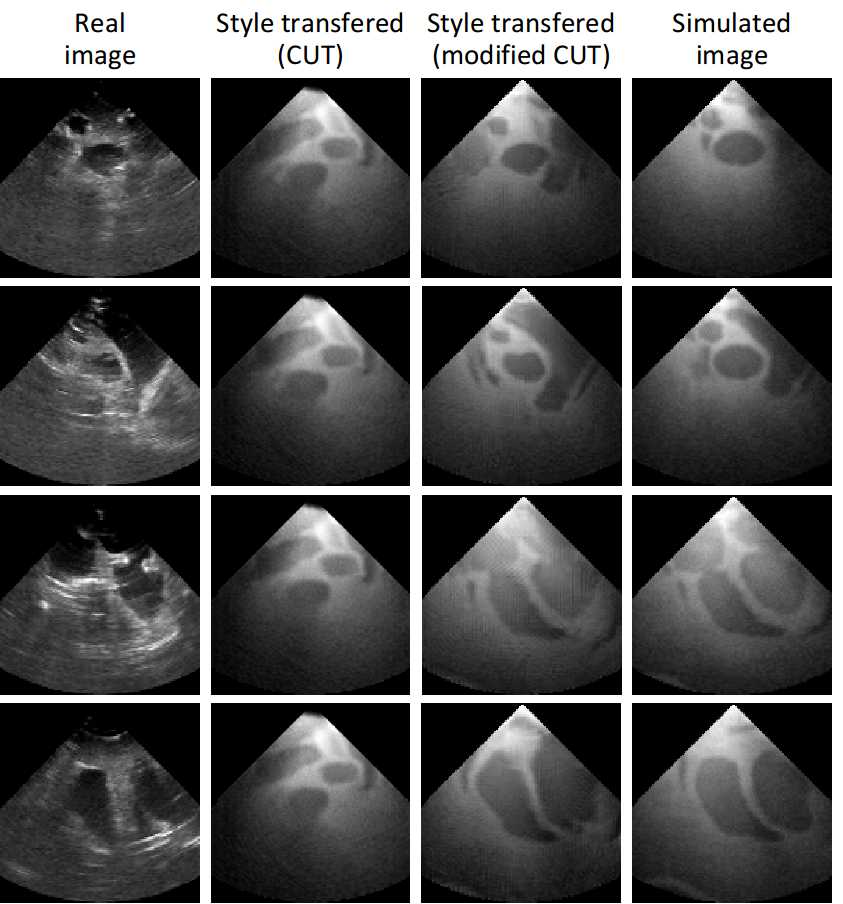}
\caption{The four rows in the first column show four real US images acquired at different views. The second column shows the style transferred images using the original CUT model, and the third column shows the style transferred images using our modified CUT model. The final column shows the simulated images corresponding to the four views.}
\label{Fig_style}
\end{figure}

We first take a look at the performance of the trained ST model based on the modified CUT method. For comparison, we also trained an original CUT model without using the SSIM loss, similar to \cite{velikova2022cactuss}.
Fig. \ref{Fig_style} shows some results obtained by the learned ST models. It can be seen from the second column that the style transferred images using the original CUT model successfully learned to change the appearance of the input real image to the simulated images (see the fourth column). However, all four real images obtained from four different views were converted into almost the same image, showing that the model failed to preserve the structural features of the original image, which are crucial for the robotic US analysis and navigation tasks that strongly rely on the structural information. In comparison, our modified CUT model shows the superiority of both taking on the simulation style and retaining the structure of the original real US images (see the third column).

\subsection{Evaluation of the Overall Performance of the Sim2Real Framework}

\begin{figure}[t]
\setlength{\abovecaptionskip}{-0.0cm}
\centering
\includegraphics[scale=1.0,angle=0,width=0.49\textwidth]{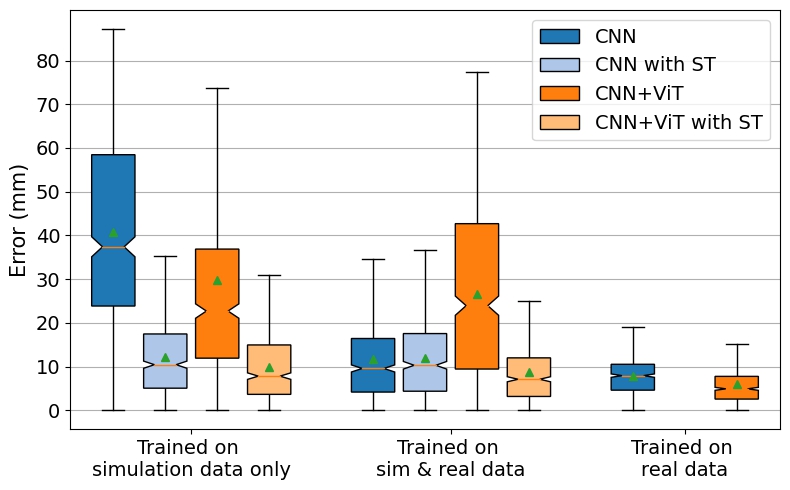}
\caption{Quantitative evaluation of the Sim2Real framework. Blue and orange colors indicate the ``CNN" and ``CNN+ViT" architectures used in the task-relevant model, respectively. Light and dark colors represent the results obtained by the model with and without using the style transfer (ST) module for preprocessing, respectively.}
\label{Fig_boxplot}
\end{figure}

First, we conducted a quantitative evaluation of the performance of the proposed Sim2Real framework. The overall localization accuracy of different configurations are reported in Table \ref{T_quantitative} and Fig. \ref{Fig_boxplot}.  The localization is considered successful if the prediction error is smaller than \SI{10}{\mm}.

We first compared the models with ``CNN" and ``CNN + ViT" structures trained using real data in a supervised fashion for the image-based localization task. As shown in Table \ref{T_quantitative} (lines 2-3) and Fig. \ref{Fig_boxplot} (last 2 columns), the ``CNN + ViT" model improved the mean localization accuracy of the ``CNN" model by 24\%, and the success rate over the 563 test data increased from 70\% to 88\%. These results show that the introduction of the self-attention mechanism can improve the generalization ability of the task-relevant model to interpret unseen US images. 

Then, we trained the task-relevant models using combined simulated data and a few real data, and tested the models on real data. As shown in Table \ref{T_quantitative} (lines 4 and 6) and Fig. \ref{Fig_boxplot}, a deterioration of was observed compared with the supervised setting, which may be due to the style differences between the simulated and real images. After applying the ST model as a preprocessing module, we found that the performance of the ``CNN + ViT" model was largely improved, with the average localization error reduced from \SI{26.53}{\mm} to \SI{8.67}{\mm} and the success rate increased from 27\% to 66\%. Meanwhile, the performance of the ``CNN" model did not change significantly before and after adding the ST module, yielding an average accuracy of about \SI{12}{\mm}.

Finally, we tested the models trained only using simulation data. It can be seen from Table \ref{T_quantitative} (lines 8 and 10) and Fig. \ref{Fig_boxplot} that the accuracy of both ``CNN" and ``CNN + ViT" models without using ST was significantly deteriorated compared with the supervised and semi-supervised settings, while the ``CNN + ViT" model showed a stronger ability to understand the similarity of content between the simulated and real images, thus largely improving the accuracy of the CNN-only model by 27\%. 
Furthermore, we applied the ST module to convert the style of the real US images, and found a remarkable improvement for the localization accuracy of the ``CNN" and ``CNN + ViT" models from \SI{40.78}{\mm} and \SI{29.81}{\mm} to \SI{12.20}{\mm} and \SI{9.90}{\mm}, respectively. Also, the success rate of the two models was increased to 48\% and 59\%, respectively, which were on par with that in the semi-supervised setting. The best result was achieved by the ``CNN + ViT" model with ST, yielding an overall localization error of 9.90 $\pm$ \SI{8.98}{\mm}, which was close to the best models trained using combined data and pure real data (with an average accuracy of \SI{8.67}{\mm} and \SI{5.93}{\mm}, respectively). The results show that the proposed ST module and CNN-ViT structure in the task-relevant model can effectively bridge the gap between the simulated and real US images, demonstrating the feasibility to deploy a US image analysis model trained only on simulated data to real-world applications without requiring labelled real data for training.

\begin{table*}[tb] \renewcommand\arraystretch{1.3} \small
\centering
\caption{Quantitative Evaluation of the Sim2Real Framework for Image-based TEE Probe Localization Task}
\resizebox{0.8\textwidth}{26 mm}{
\begin{tabular}{m{2.8cm}<{\raggedright}m{3cm}<{\raggedright}m{2cm}<{\raggedright}m{3cm}<{\centering}m{2cm}<{\centering}}
\Xhline{1pt}
Training data  & Task-relevant model  &Style transfer&	Error (\si{mm}) & Success rate \\
\Xhline{1pt}
\multirow{2}{*}{Real data} & CNN	&-	&7.77	$\pm$ 4.08	& 70\% \\
\cline{2-5}
& CNN + ViT & -&5.93 $\pm$	4.94 &	88\%\\
\hline
\multirow{4}{*}{Combined data} &\multirow{2}{*}{CNN}	&Not used	&11.70	$\pm$ 9.68&	52\%\\
\cline{3-5}
&&Used &12.03	$\pm$ 9.29	&49\%\\
\cline{2-5}
&\multirow{2}{*}{CNN + ViT} &	Not used&	26.53 $\pm$ 19.21	&27\%\\
\cline{3-5}
&&Used	&8.67	$\pm$ 7.88	  &66\% \\
\hline
\multirow{4}{*}{Simulation data} & \multirow{2}{*}{CNN} & 	Not used&	40.78 	$\pm$ 23.68 &	13\% \\
\cline{3-5}
&&Used & 12.20	$\pm$ 8.86 &	48\%\\
\cline{2-5}
&\multirow{2}{*}{CNN + ViT} &	Not used	&29.81 $\pm$ 26.95 & 	21\%\\
\cline{3-5}
&&Used & 9.90	$\pm$ 8.98 &	59\%\\
\Xhline{1pt}
\end{tabular}}
\label{T_quantitative}
\end{table*}

\begin{figure}[t]
\setlength{\abovecaptionskip}{-0.0cm}
\centering
\includegraphics[scale=1.0,angle=0,width=0.3\textwidth]{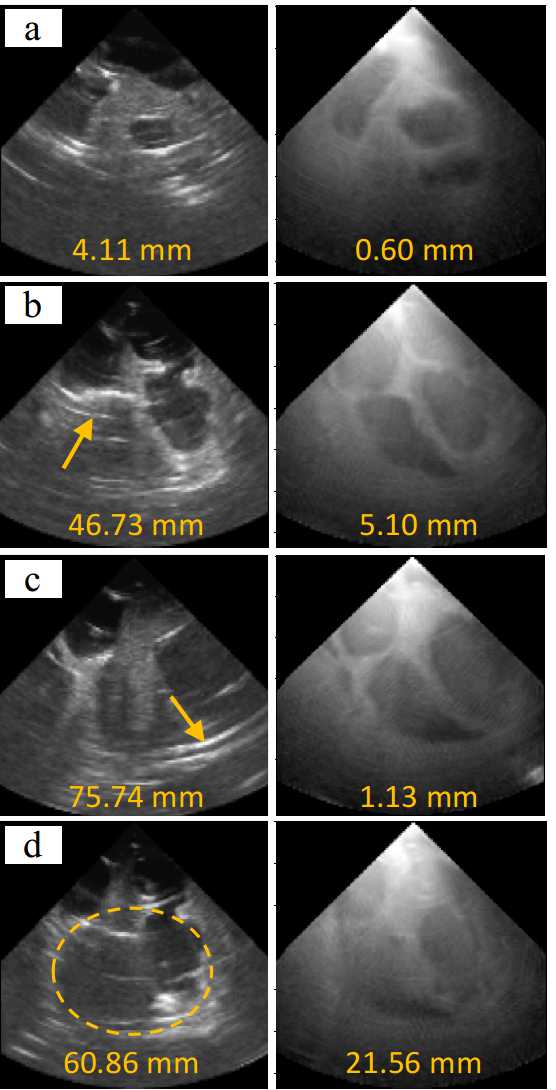}
\caption{Qualitative comparison of the methods with and without using the ST module. The task-relevant model was trained only using simulation data and tested on real US data. The errors are marked using yellow color. The arrows and dashed circle are used to highligh the US artifacts in the real US images.}
\label{Fig_qualitative}
\end{figure}

To further assess the performance of the Sim2Real framework, we performed a qualitative analysis by taking a closer look at the results of the framework when adopting the ``CNN + ViT" as the task-relevant model and training it using only the simulation data to see how the ST module helps bridge the Sim2Real gap and thus improving the localization performance. Fig. \ref{Fig_qualitative} illustrates 4 testing images with the style transferred ones using our proposed ST module. The errors are marked in the images. 
Fig. \ref{Fig_qualitative}(a) shows a situation where the application of the ST module made little difference in the localization results. This may be because the original real image is sufficiently clear and high-contrast, with features suitable for localization. Fig. \ref{Fig_qualitative}(b-c) illustrate situations where employing the ST module significantly improves the localization performance. It can be observed from the two real images that there exist some US artifacts appearing as bright lines in the images, which may represent a false portrayal of anatomy and may degrade the accuracy of the image analysis model. In comparison, the ST module converted the image into the simulation style, which mitigated the impact of the US artifacts and helped the task-relevant model better focus on the true anatomical information in the image.
Fig. \ref{Fig_qualitative}(d)  shows a situation where the localization performance of both methods are equally poor no matter whether the ST module is used or not. It can be seen that a large shadowing artifact appears in the real US image, which may be caused by a partial loss of contact between the probe and the tissue. Since the key information representing the anatomical structure in the original image is lost, although the use of style enhancement can improve the performance of the model to a certain extent, the final localization accuracy is still not satisfactory. Nevertheless, performing diagnosis based on such US images with poor quality is also challenging for expert sonographers. In view of this, force-control-based strategies should be employed to ensure acoustic coupling during robotic US acquisitions, while this is beyond the scope of this paper.

\section{Conclusions}

In this paper, we present a novel Sim2Real framework based on the combination of style transfer and self-attention-enhanced task-relevant models for end-to-end robotic US image analysis tasks. 
The style transfer module is designed based on an unsupervised contrastive learning approach, and the task-relevant model is designed as a concatenated CNN-ViT structure to capture long-range dependencies in data for better generalization capabilities.
We validated the proposed framework an image regression task to predict the TEE probe position in the esophagus based on the acquired US images with a model trained only using simulation data.
We show that without using labelled real US data for training, our proposed method that applies the self-attention mechanism in the task-relevant model and uses a style transfer module for preprocessing can effectively bridge the gap between simulated and real images, realizing comparably satisfactory performance compared with semi-supervised and fully supervised methods. 
The effectiveness of our previously proposed CT-based US image simulation method is also indirectly confirmed. 




\addtolength{\textheight}{-6cm}   

\bibliographystyle{IEEEtran}   
\bibliography{root}

\end{document}